\documentclass{article}

\usepackage{arxiv}

\usepackage[utf8]{inputenc} % allow utf-8 input
\usepackage[T1]{fontenc}    % use 8-bit T1 fonts
\usepackage{hyperref}       % hyperlinks
\usepackage{url}            % simple URL typesetting
\usepackage{booktabs}       % professional-quality tables
\usepackage{amsfonts}       % blackboard math symbols
\usepackage{nicefrac}       % compact symbols for 1/2, etc.
\usepackage{microtype}      % microtypography
\usepackage{lipsum}		% Can be removed after putting your text content
\usepackage{graphicx}
\usepackage{natbib}
\usepackage{doi}
\RequirePackage{algorithm}
\RequirePackage{algorithmic}
\RequirePackage{algorithm}
\RequirePackage{algorithmic}
\usepackage{multicol}
\usepackage{float}
\usepackage{setspace}
\usepackage{hyperref}
\usepackage{url}
\usepackage{lipsum}
\usepackage{amsthm}
\usepackage{booktabs} 
\usepackage{amsfonts,bm}
\usepackage{subcaption}
\usepackage{caption}
\usepackage{tabularx}
\usepackage{xspace}
\usepackage{booktabs} % commands to create good-looking tables
\usepackage{tikz} % nice language for creating drawings and diagrams
\usepackage{times}
\usepackage{epsfig}
\usepackage{graphicx}
\usepackage{amsmath}
\usepackage{amssymb}

\newtheorem{definition}{Definition}
\newtheorem{theorem}{Theorem}
\newcounter{lemma}[theorem]
\newtheorem{lemma}[theorem]{Lemma}

\newcommand{\argminA}{\arg\,\min}
\newcommand\numberthis{\addtocounter{equation}{1}\tag{\theequation}}

\title{Learning Generative Prior with Latent Space Sparsity Constraints}

\author{ {\hspace{1mm}Vinayak Killedar} \\
	Department of Electrical Engineering\\
	Indian Institute of Science\\
	Bangalore-560012, India \\
	\texttt{vinayakk1@iisc.ac.in} \\
	\And
	{\hspace{1mm}Praveen Kumar Pokala} \\
	Department of Electrical Engineering\\
	Indian Institute of Science\\
	Bangalore-560012, India \\
	\texttt{praveenkumar@iisc.ac.in} \\
	\And
	{\hspace{1mm}Chandra Sekhar Seelamantula} \\
	Department of Electrical Engineering\\
	Indian Institute of Science\\
	Bangalore-560012, India \\
	\texttt{chandra.sekhar@ieee.org} \\

}

\begin{document}
\maketitle
\begin{abstract}
We address the problem of compressed sensing using a deep generative prior model and consider both linear and learned nonlinear sensing mechanisms, where the nonlinear one involves either a fully connected neural network or a convolutional neural network. Recently, it has been argued that the distribution of natural images do not lie in a single manifold but rather lie in a union of several submanifolds. We propose a sparsity-driven latent space sampling (SDLSS) framework and develop a proximal meta-learning (PML) algorithm to enforce sparsity in the latent space. SDLSS allows the range-space of the generator to be considered as a union-of-submanifolds. We also derive the sample complexity bounds within the SDLSS framework for the linear measurement model. The results demonstrate that for a higher degree of compression, the SDLSS method is more efficient than the state-of-the-art method. We first consider a comparison between linear and nonlinear sensing mechanisms on Fashion-MNIST dataset and show that the learned nonlinear version is superior to the linear one. Subsequent comparisons with the deep compressive sensing (DCS) framework proposed in the literature are reported. We also consider the effect of the dimension of the latent space and the sparsity factor in validating the SDLSS framework. Performance quantification is carried out by employing three objective metrics: peak signal-to-noise ratio (PSNR), structural similarity index metric (SSIM), and reconstruction error (RE). 
% Experimental validation demonstrates that the proposed SDLSS framework is superior to the DCS framework. 
\end{abstract}

\section{Introduction}\label{sec:intro}
The goal of compressed sensing (CS) is to reconstruct a signal $\boldsymbol{x}  \in \mathbb{R}^{n}$ from a set of $m$ linear or nonlinear measurements $\boldsymbol{y}  \in \mathbb{R}^{m}$, where $m \ll n$. The standard CS formulation considers a linear sensing operator and relies on the sparsity of the ground-truth vector $\boldsymbol{x}$. The problem is, in general, ill-posed and has infinite solutions. One could narrow down the search space of solutions by imposing certain structural assumptions on the unknown signal $\boldsymbol{x}$ and sparsity is one such prior in the standard CS setup. Consider the generic problem setting
\begin{equation}\label{eq1}
\boldsymbol{y} = \mathcal{A}(\boldsymbol{x}),
\end{equation}
where the sensing operator $\mathcal{A}$ may be linear or nonlinear. Developing a recovery algorithm involves imposing a desired structure on $\boldsymbol{x}$ and designing an efficient algorithm to estimate $\boldsymbol{x}$ that is consistent with the measurements $\boldsymbol{y}$.

\indent In standard CS, considering a noise-free measurement, the optimization problem takes the following form:
\begin{equation}\label{eq2}
    \min_{\boldsymbol{\alpha}} \quad\Vert {\boldsymbol{\alpha}} \Vert_0\quad \text{s.t.} \quad \boldsymbol{y} = \mathbf{A}\boldsymbol{\Phi \alpha} \quad \text{and} \quad \boldsymbol{x} = \boldsymbol{\Phi}\boldsymbol{\alpha},
\end{equation}
where $\mathbf{A} \in \mathbb{R}^{m \times n}$ is the sensing matrix, $\boldsymbol{\alpha}$ is the sparse code, and $\boldsymbol{x}$ is assumed to be $s$-sparse in an overcomplete dictionary $\boldsymbol{\Phi}$.
Recovery of $\boldsymbol{\alpha}$ in~\eqref{eq2} is guaranteed with a high probability if $\mathbf{A}$ satisfies the Restricted Isometric Property (RIP)~\citep{rip} and $\boldsymbol{x}$ lies in the range-space of $\boldsymbol{\Phi}$. Sparsity has enabled the design of acquisition strategies with a reduced sample complexity. Compressed Sensing has directly impacted several imaging modalities, magnetic resonance imaging (MRI)~\citep{lustig2007sparse,jacob2020computational} being a prime example. The success of CS has also been observed in several computer vision problems ~\citep{wang2016sparse}.

\indent In this paper, we consider generative model based compressive sensing (GMCS) with a noiseless measurements. More precisely, we pursue the following objective:
\begin{equation}\label{eq3a}
    \min_{\boldsymbol{z}} \quad\Vert \boldsymbol{z} \Vert_0\quad \text{s.t.} \quad \boldsymbol{y} = \mathbf{A} \mathbf{G}_{\boldsymbol{\theta}}(\boldsymbol{z}), \quad \boldsymbol{x} = \mathbf{G}_{\boldsymbol{\theta}}(\boldsymbol{z}),
\end{equation}
where $\mathbf{A} \in \mathbb{R}^{m \times n}$ is the sensing matrix with Gaussian i.i.d. entries $\mathbf{A}_{i,j}\sim \mathcal{N}(0,\frac{1}{m})$ , $\mathbf{G}_{\boldsymbol{\theta}}: \mathbb{R}^{k} \rightarrow \mathbb{R}^{n}$ is the generator model, and $\boldsymbol{z} \in  \mathbb{R}^{k}$ is the latent space representation sampled from the standard normal distribution $\mathcal{N}(0,I)$. The problem stated in~\eqref{eq3a} is solvable if $\mathbf{A}$ satisfies the Set-Restricted Eigenvalue Condition ($\mathop{S}$-REC) \citep{Bora} (also defined subsequently, in Definition~\ref{SREC}) and  $\boldsymbol{x}$ lies in the range-space of the generator $\mathbf{G}_{\boldsymbol{\theta}}$.

\section{Prior Art}
\indent Several CS reconstruction techniques~\citep{ribes2008linear,borgerding2017amp,zhang2018ista, pokala2019firmnet,  kamilov2016learning} have been developed over the past decade by exploiting the  sparsity prior either in the signal domain or in a suitable transform domain. Traditional CS reconstruction is posed as a sparsity-regularized optimization problem and solved iteratively~\citep{beck2009fast,daubechies2004iterative}.
LASSO~\citep{lasso} solves a tractable convex optimization problem for sparse recovery. Most iterative techniques enjoy the advantages of theoretical analysis and  convergence properties, but they are typically computationally intensive and one needs to know the optimal sparsifying transform a priori. Learning based CS reconstruction techniques~\citep{zhang2018ista, pokala2019firmnet, kamilov2016learning,mukherjee2017dnns,jawali2020cornet} are inspired by iterative algorithms and result in superior recovery performance.

\indent In generative model based compressed sensing (GMCS)~\citep{Bora}, a pretrained network $\mathbf{G}_{\boldsymbol{\theta}}$ acts as a generative prior as opposed to the sparsity based prior in standard CS. The foundation for GMCS lies in the {\it universal approximation theorem} (UAT)~\citep{uat}, which states that, under certain conditions, a neural network can approximate any regular function with arbitrary precision.
Most commonly used generative models are variational auto-encoders (VAE)~\citep{kingma2014autoencoding} and generative adversarial networks (GAN)~\citep{Good}.
The pretrained generative model captures a low-dimensional representation of $\boldsymbol{x}$ based on the data that it has seen during training. GMCS has been shown to outperform LASSO \citep{lasso} when the sensing rates are low~\citep{Bora}. The success of GMCS~\citep{Bora} lies in the signal $\boldsymbol{x}$ belonging to the range-space of the pretrained generator,  $\mathbf{G}_{\boldsymbol{\theta}}: \mathbb{R}^{k} \rightarrow \mathbb{R}^{n}$ such that $\boldsymbol{x} = \mathbf{G}_{\boldsymbol{\theta}}(\boldsymbol{z})$, where $\boldsymbol{z} \in \mathbb{R}^{k}$  and $\boldsymbol{\theta}$ denotes the latent variable and the parameter of the generator, respectively. If the range-space condition is violated, then there would be an error in the reconstruction \citep{dhar2018modeling,Bora}. Also, the reconstruction process is slow because the optimal latent space is discovered through gradient-descent (GD) optimization, which requires several restarts~\citep{Yan}.

\indent Several methods have been proposed to improve the representation accuracy and reduce the reconstruction error.
\citep{dhar2018modeling} introduced a sparse deviation $\boldsymbol{v}$ in the range-space of the generator $\mathbf{G}_{\boldsymbol{\theta}}(\boldsymbol{z})$, i.e., they considered the range-space to be $\mathbf{G}_{\boldsymbol{\theta}}(\boldsymbol{z}) + \boldsymbol{v}$ and solved an alternating minimization problem with respect to $\boldsymbol{z}$ and $\boldsymbol{v}$ \citep{dhar2018modeling}.
To overcome the limited expressiveness of the latent variable due to reduced dimensionality, ~\citep{mganprior} proposed a multi-code GAN prior called as mGANprior, which uses multiple latent variables to generate multiple feature maps. The feature maps are then combined based on relative channel importance and fed as input to the generator. In flow-based and invertible multiscale generative models~\citep{invetgan,varuninvetgan}, the generator model $\mathbf{G}_{\boldsymbol{\theta}}: \mathbb{R}^{n} \rightarrow \mathbb{R}^{n}$ has zero representation error, because the dimension of the latent space and the input space are the same. The generator is an invertible network and images that are not in the distribution of the training set can also be recovered.
Deep Image Prior~\citep{deepimageprior} and Deep Decoder~\citep{deepdecoder} are a different class of approaches that use an untrained neural network prior (a neural network with random weights) to capture the low-level signal statistics. During testing, the parameters of the neural network are optimized to best fit the measurements.
Image adaptive GAN (IAGAN)~\citep{imageadaptive} reduces the representation error by simultaneously optimizing the parameters of the pretrained network and the latent variable during testing. Shah and Hegde~\citep{chinmay} proposed a projected gradient-descent (PGD) algorithm with theoretical guarantees to solve the linear inverse problem with a generative prior by optimizing the signal space and projecting back on to the latent space. To reconstruct high-resolution natural images, a convolutional manifold structure is imposed on the high-dimensional latent space (i.e., the dimension of the latent space is more than that of the signal to be reconstructed) in the latent convolutional model~\citep{latentconvm} by using an additional convolutional neural network. The reconstruction process in the preceding methods is slow because during inference, a pretrained generator network is used and the gradient-descent (GD) optimization is performed on the latent space with several restarts.

\indent Deep compressed sensing (DCS) framework~\citep{Yan} demonstrated that the joint training of the generator and the optimization of the latent space via meta-learning~\citep{finn} leads to a faster and accurate reconstruction in comparison with~\citep{Bora,dhar2018modeling}.  \citep{infogan} enforced a structure on the latent space $\boldsymbol{z}$ through InfoGAN, which resulted in faster CS recovery \citep{Romberg}.

\indent A sample complexity bound for compressive sensing using generative model with ReLU activation was proposed by~\citep{Bora}. ~\citep{scarlet} derived an algorithm-independent lower bound on the sample complexity under group sparsity prior on the signal to be reconstructed. ~\citep{scnlm} presented sample complexity bounds for nonlinear measurement models with heavy-tailed noise. ~\citep{VAEIVp} showed that if the number of measurements $m$ is larger than the dimension $k$ of the input to the generative model, then asymptotically, solving inverse problems using a variational autoecndoer (VAE) has almost zero recovery loss. ~\citep{hand} developed global guarantees for CS with generative modelling and showed that gradient-descent optimization can reach global minima with high probability when the objective function given by empirical risk does not have spurious stationary points.

% \begin{figure}[t]
%   \centering
%   \begin{subfigure}{.35\textwidth}
%     \centering\includegraphics[width=\textwidth]{figures/celeba_meas_spr500new.pdf}
%   \end{subfigure}
%   \caption{Illustration of the advantages of the proposed SDLSS framework over DCS on CelebA dataset. The latent space dimension  $k = 4096$. Each row  corresponds to a different number of measurements $m$. The sparsity factor $s$ in SDLSS is set to $500$.}
%   \label{RE_Celeba_meas}
% \end{figure}

\subsection{Motivation and Contribution}
GMCS considers a single generator to model the distribution of signal in the solution space. 
However, signals such as natural images do not lie in a single manifold but rather lie in a union-of-submanifolds that have minimal overlap~\citep{NIPS2018_7964}. A single generator cannot correctly model such a distribution \citep{NIPS2018_7964}. \citep{NIPS2018_7964} proposed a collection of generators to model the data distribution that lies in a union of disconnected manifolds and suggested that the disconnectedness could be factored into the latent space to improve the representation. \citep{maskAAE} hypothesized that the dimensionality of the latent space in an autoencoder model has a critical effect on the generated image.
Flow-based and invertible multiscale generative models~\citep{invetgan,varuninvetgan} showed that images that are not in the distribution of the training set can be reconstructed by making the dimensionality of the latent space equal to the ambient dimension. A better representation of the signal $\boldsymbol{x}$ by introducing sparsity in the latent space through copula transformation was shown by~\citep{copula}.
Motivated by the success of the prior works, we propose a novel sparsity-driven latent space sampling (SDLSS) framework to improve the reconstruction capability. The objective is to reduce the representation error by considering a latent space dimension that is equal to that of the output dimension of the generator but with sparsity being enforced in the latent space while training the generator. The recovery with sparsity constraint is achieved with the help of a {\it proximal meta-learning} algorithm. We demonstrate the efficacy of the proposed approach considering several standard datasets such as Fashion-MNIST, CIFAR-$10$, and celebA. Further, we derive the sample complexity bounds within the SDLSS framework for linear measurement model considering a piecewise-linear activation. 
% Figure~\ref{RE_Celeba_meas} gives a feel for the results that could be obtained with SDLSS vis-\`a-vis DCS, particularly at higher compression rates.

\section{Signal Models}
\begin{figure}[t]
\centering
\begin{subfigure}{.3\textwidth}
    \centering\includegraphics[width=\textwidth]{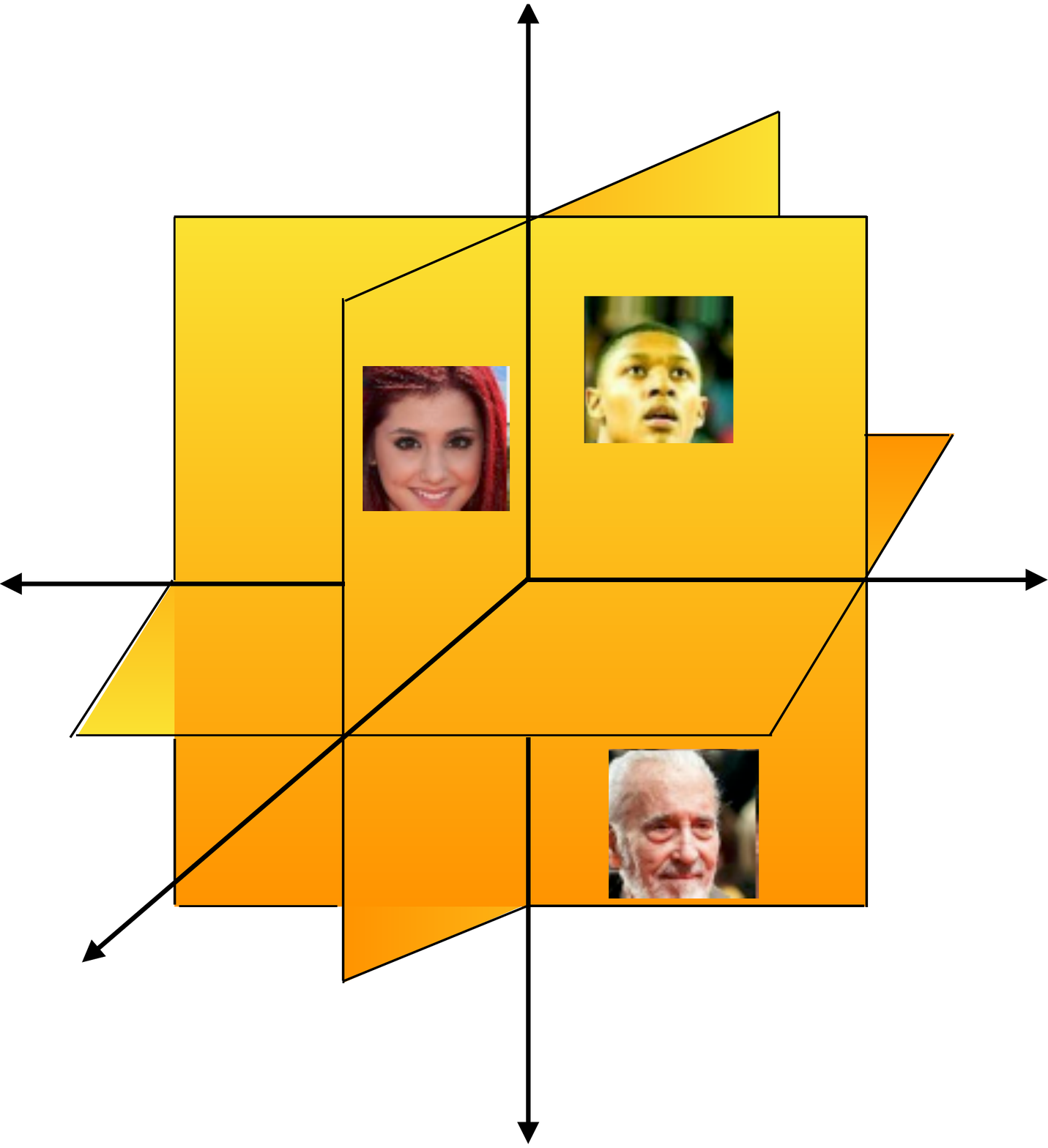}
    \caption{}
  \end{subfigure}
  ~
\begin{subfigure}{.3\textwidth}
    \centering\includegraphics[width=\textwidth]{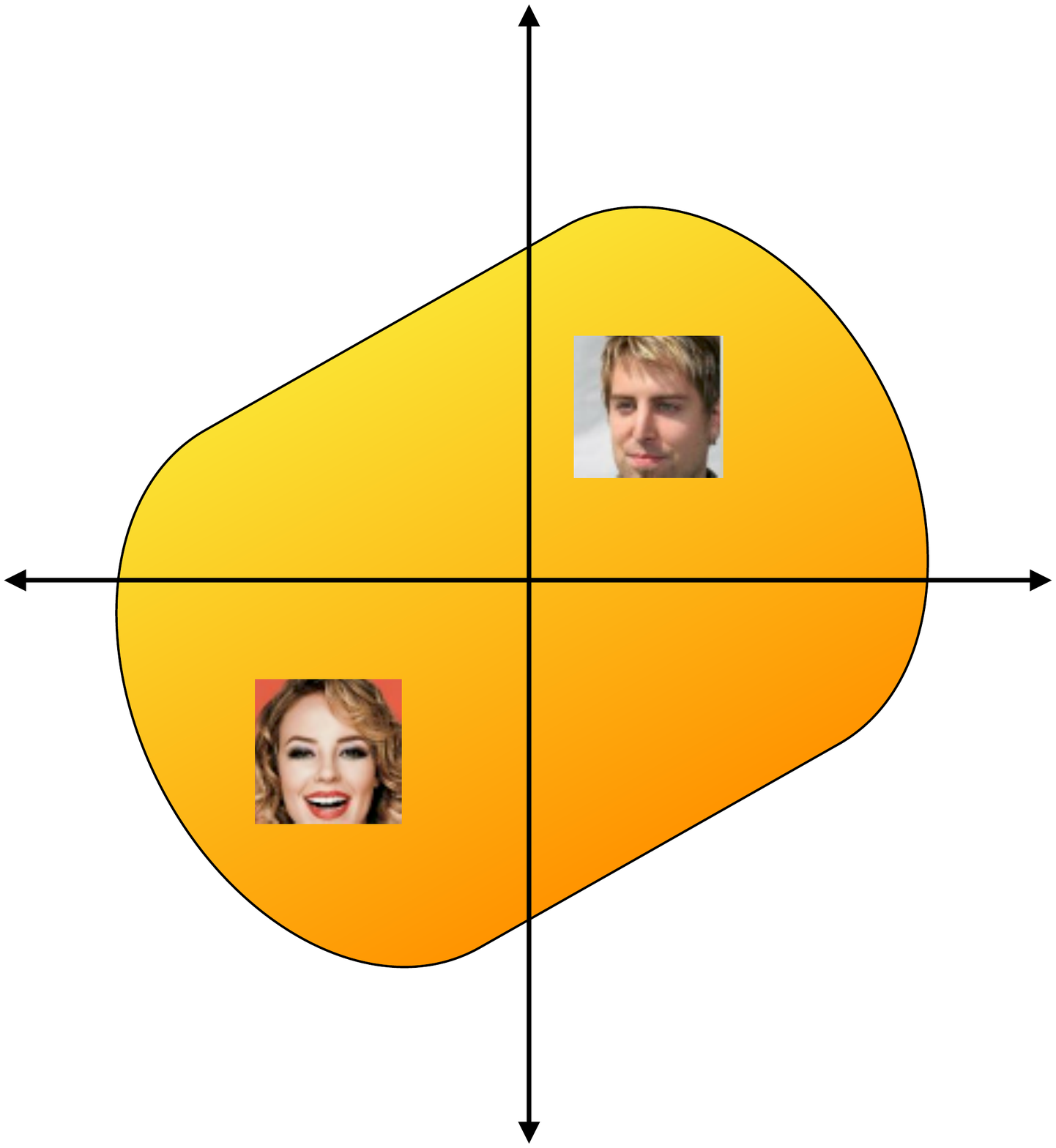}
    \caption{}
  \end{subfigure}
  ~
\begin{subfigure}{.3\textwidth}
    \centering\includegraphics[width=\textwidth]{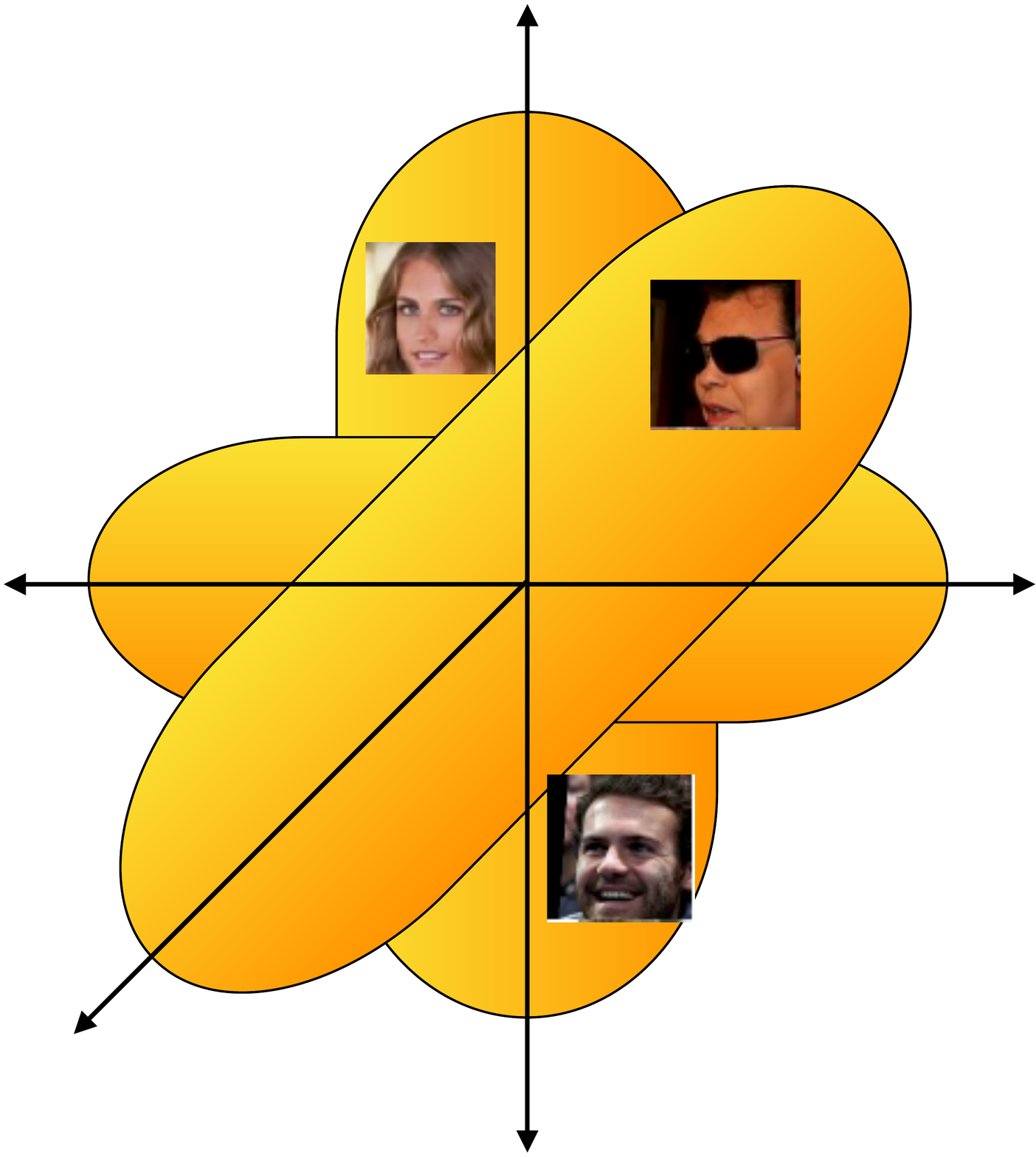}
    \caption{}
  \end{subfigure}
\caption{Signal models: (a) Union-of-subspaces, (b) Generative model, and (c) Union-of-submanifolds.}
\label{signalmodel}
\end{figure}

In this section, we review the union-of-subspaces model introduced in~\citep{unionss} based on the sparsity prior and correspondingly propose a union-of-submanifolds models for the sparsity driven generative prior. The various signal models are illustrated in Fig.~\ref{signalmodel}.
\subsection{Union of Subspaces}
Consider the standard CS problem formulated as follows. Given
\begin{equation}\label{eq3}
    \boldsymbol{y} = \mathbf{A}\boldsymbol{x},
\end{equation}
where $\mathbf{A} \in \mathbb{R}^{m \times n}$ is the sensing matrix with $m \ll n$, the goal is to recover $\boldsymbol{x}$ from $\boldsymbol{y}$ under the prior that $\boldsymbol{x}$ lies in a union-of-subspaces, given by
\begin{equation}\label{eq4}
    \mathcal{U} = \bigcup_{i} \mathcal{V}_i,
\end{equation}
where each $\mathcal{V}_i$ is a subspace in itself.
The sparsity assumption on $\boldsymbol{x}$ implies that it lies in one of the unknown subspaces  $\mathcal{V}_i$, which is not known a priori. However, if $\boldsymbol{x}$ is sparse in the transform domain, an additional structure is assumed on $\mathcal{V}_i$, which is given below.
\begin{equation}\label{eq5}
    \mathcal{V}_i = \bigoplus_{\substack{{j \in \mathcal{J}_i}\\|\mathcal{J}_i| = k}} \mathcal{W}_j,
\end{equation}
where $\left\{\mathcal{W}_j\right\}_{j=1}^{n}$ are a given set of disjoint subspaces, $\bigoplus$ denotes direct-sum, $|\cdot|$ stands for cardinality of a set, and $\mathcal{J}_i \subseteq \{1, \ldots, n\}$. 
Therefore, there are $\binom{n}{k}$ number of  $\mathcal{V}_i$ subspaces comprising the union.
A special case of the union-of-subspaces model is the standard compressive sensing problem in which $\boldsymbol{x} \in \mathbb{R}^{n}$ has a sparse representation in a given basis $\boldsymbol{\Phi} \in \mathbb{R}^{n \times n}$ given as:
\begin{equation}\label{eq6}
    \boldsymbol{x} = \boldsymbol{\Phi \alpha},
\end{equation}
where $\boldsymbol{\alpha}$ is a $k$-sparse vector.
Thus, choosing $\mathcal{W}_i$ to be the space spanned by the $i^\text{th}$ column of $\boldsymbol{\Phi}$ fits well with the union-of-subspaces model.

\subsection{Union of Submanifolds}
\indent Manifold based signal recovery from compressed measurement has been proposed by~\citep{eftekhari2015new} and ~\citep{hegde2012signal}. 
In this work, we assume $s$-sparse high dimensional latent variable $\boldsymbol{z} \in \mathbb{R}^{k}$ as input to the generator. 
The generator model is given as
\begin{equation}\label{eq8}
    \boldsymbol{x} = \mathbf{G}_{\boldsymbol{\theta}}(\boldsymbol{z}) \quad \text{s.t.} \quad \Vert\boldsymbol{z}\Vert_0 \leq s.
\end{equation}
Analogous to the classical CS where the reconstructed signal $\boldsymbol{x} \in \mathbb{R}^{n}$ obeys the union-of-subspaces model, which matches with the range-space of an overcomplete dictionary, in the SDLSS framework, the range-space of the generator network is assumed to match closely with the union-of-submanifolds model that contains the signal. Let the subspace spanned by a given $s$-sparse latent variable $\boldsymbol{z}$ be denoted as $\mathcal{W}_i$.
The generator $\mathbf{G}_{\boldsymbol{\theta}}: \mathbb{R}^{k} \rightarrow \mathbb{R}^{n}$ could be interpreted as a non-linear dictionary that maps each subspace $\mathcal{W}_i$ to a corresponding submanifold $\mathcal{S}_i \in \mathbb{R}^{n}$.

The sparsity ${s}$ assumption on the input latent variable divides the latent space into $\binom{k}{s}$ subspaces $\left\{\mathcal{W}_i\right\}$, such that the generator model $\mathbf{G}_{\boldsymbol{\theta}}$ transforms each subspace $\mathcal{W}_i$ to the corresponding submanifold $\left\{\mathcal{S}_i\right\}$. Thus, the range-space of the generator $\mathbf{G}_{\boldsymbol{\theta}}: \mathbb{R}^{k} \rightarrow \mathbb{R}^{n}$ is composed of a union of $\binom{k}{s}$ number of submanifolds $\left\{\mathcal{S}_i\right\}$, i.e.,
\begin{equation}\label{eq9}
     \mathcal{S}_{s,\mathbf{G}_{\boldsymbol{\theta}}} = \bigcup_{i} \mathcal{S}_i,
\end{equation}
where $\mathcal{S}_i$ is the submanifold generated by $\mathbf{G}_{\boldsymbol{\theta}}(\boldsymbol{z})$, with $\Vert \boldsymbol{z} \Vert_0 \leq s$ and $\mathcal{S}_{s,\mathbf{G}_{\boldsymbol{\theta}}}$ denotes the range-space of generator $\mathbf{G}_{\boldsymbol{\theta}}$ with $s$-sparse latent representation $\boldsymbol{z}$.

\begin{definition}\label{SREC}
\citep{Bora} Let $\mathcal{S} \subseteq \mathbb{R}^{n}$, $\gamma > 0$, and $\delta \geq 0$. Matrix $\mathbf{A} \in \mathbb{R}^{m \times n}$ is said to satisfy the $\mathop{S}$-REC$(\mathcal{S},\gamma,\delta)$ if $\forall \boldsymbol{x}_{1}, \boldsymbol{x}_{2} \in \mathcal{S},$
\begin{equation}\label{eq11}
    \Vert \mathbf{A}(\boldsymbol{x}_{1} - \boldsymbol{x}_{2}) \Vert_2 \geq \gamma \Vert \boldsymbol{x}_{1} - \boldsymbol{x}_{2} \Vert_2 - \delta.
\end{equation}
\end{definition}
\noindent Classical CS algorithms impose REC~\citep{REC} on the measurement matrix $\mathbf{A}$ to recover an $s$-sparse signal. In GMCS recovery, $\mathop{S}$-REC generalizes REC to a set $\mathcal{S}$ of vectors that are in the range-space of the generator.\\

\section{Sparsity-Driven Latent Space Sampling}
The sparsity based prior in CS is a weak one and results in a large reconstruction error when the number of measurements is not adequate (high compression ratios). On the contrary, generative models with a fixed dimension $k$ for the input latent space $\mathcal{Z}$ assume a stronger prior that the reconstructed signal should lie in the range-space of the generator $\mathbf{G}_{\boldsymbol{\theta}}$.
Natural signals do not lie in a single manifold modeled by the range-space of the generator. Instead, they could be modeled as belong to a union of disconnected submanifolds~\citep{NIPS2018_7964}. Since a single generator cannot accurately model a distribution that spans several disconnected manifolds, one resorts to alternatives such as employing a mixture model (mixture of Gaussians, for instance~\citep{deligan}) or a hybrid model that models certain dimensions as discrete~\citep{infogan}. The other approach is to employ a collection of multiple generators~\citep{NIPS2018_7964}.

\indent In this paper, we propose a hybrid approach -- we combine the advantages of sparsity driven modeling along with generative modeling. More precisely, the latent space dimension is comparable to that of the input to the measurement model, but with sparsity enforced to introduce discreteness in the representation space. The combination is capable of parsimoniously modeling a union of disconnected submanifolds. 
Thus, SDLSS framework allows the signal $\boldsymbol{x}$ to lie in the union-of-submanifolds $\mathcal{S}_{s,\mathbf{G}_{\boldsymbol{\theta}}}$ spanned by the generator network $\mathbf{G}_{\boldsymbol{\theta}}$. We consider DCS~\citep{Yan} framework with a linear and learned nonlinear measurement operator (implemented using a network) and solve the following optimization problem:
\begin{align*}
    \min_{\boldsymbol{z}, \boldsymbol{\theta}} & \Vert \boldsymbol{z} \Vert_{0},
     \quad \text{s.t.}\quad \Vert \boldsymbol{y} - \mathbf{A} \mathbf{G}_{\boldsymbol{\theta}}(\boldsymbol{z}) \Vert_2 \leq \epsilon \, \, \text{and} \\ &  
    \Vert \mathbf{A}(\boldsymbol{x}_{1} - \boldsymbol{x}_{2}) \Vert_2 \geq \gamma \Vert \boldsymbol{x}_{1} - \boldsymbol{x}_{2} \Vert_2 - \delta \numberthis \label{SDLSS},
\end{align*}
where $\delta \geq 0$ and $\gamma > 0$ are user-defined parameters for matrix $\mathbf{A}$ to satisfy $\mathop{S}$-REC, $\boldsymbol{z} \in \mathbb{R}^{k}$ is the latent space with sparsity at most $s$, and ${\boldsymbol{\theta}}$ is the parameter of the generator network $\mathbf{G}_{\boldsymbol{\theta}}$.
Since $\boldsymbol{z}$ is assumed to be $s$-sparse, the $\ell_0$ pseudo-norm is implemented by means of a hard-thresholding operator that retains the $s$ largest magnitudes in $\boldsymbol{z}$ and sets the rest to zero. 
Thus, the proposed methods introduces sparsity in the continuous latent space.
% The measurement network $\mathbf{A}_{\boldsymbol{\phi}}$ is trained to satisfy the set-restricted eigenvalue condition ($\mathop{S}$-REC) given in Definition~\ref{SREC}. 
The optimization cost corresponding to the problem given in Eq.~\eqref{SDLSS} is given as follows:
\begin{equation}
        \min_{\boldsymbol{z}, \boldsymbol{\theta}}\, \left(\mathcal{L}_{\mathbf{G}} + \mathcal{L}_{\mathbf{A}}\right), \label{EQ:LA}
\end{equation}
where
\begin{align*}
    \mathcal{L}_{\mathbf{G}} & = \mathop{\mathbb{E}}_{\boldsymbol{z}}\left\{\Vert \boldsymbol{y} - \mathbf{A}\mathbf{G}_{\boldsymbol{\theta}}(\boldsymbol{z}) \Vert_2 +  \Vert\boldsymbol{z}\Vert_0\right\},\quad\text{and}\nonumber\\
    \mathcal{L}_{\mathbf{A}} & = \mathop{\mathbb{E}}_{\boldsymbol{x}_{1},\boldsymbol{x}_{2}}\{{\Vert \mathbf{A}(\boldsymbol{x}_1-\boldsymbol{x}_2)\Vert}_2 + \delta - \gamma{\Vert \boldsymbol{x}_1-\boldsymbol{x}_2 \Vert}_2\}.\nonumber
\end{align*}
The generator model $\mathbf{G}_{\mathbf{\theta}}$ is trained jointly along with latent space optimization using proximal meta learning (PML) to minimise the expected measurement loss $\mathcal{L}_{\mathbf{G}} $ and the $\mathop{S}$-REC loss  $\mathcal{L}_{\mathbf{A}}$.
The signals $\boldsymbol{x}_1$ and $\boldsymbol{x}_2$ in $\mathop{S}$-REC loss function $\mathcal{L}_{\mathbf{A}}$ are sampled from the true data distribution $p_{\text{data}}(\boldsymbol{x})$ and the generator network $\mathbf{G}_{\boldsymbol{\theta}}(\boldsymbol{z})$, respectively, and the latent variable $\boldsymbol{z}$ is sampled from the standard normal distribution $\mathcal{N}(0,I)$.
The loss function in~\eqref{EQ:LA} is defined to satisfy the $\mathop{S}$-REC defined in~$\eqref{eq11}$ and minimize the reconstruction error.
The reconstructed signal is given as $\boldsymbol{\hat{x}} = \mathbf{G}_{\boldsymbol{\theta}}(\boldsymbol{\hat{z}})$, where $\boldsymbol{\hat{z}}$ is the optimized latent space variable.

\section{Sample Complexity Results for Linear Model}
\noindent We review the lemmas provided by~\citep{Bora} and~\citep{gajjar2020subspace}, and use them to derive the sample complexity bound for the union-of-submanifolds model considering a linear measurement model.
% \begin{definition}\label{SREC}
% (Bora \it {et al.}~\citep{Bora}) Let $\mathcal{S} \subseteq \mathbb{R}^{n}$, $\gamma > 0$, and $\delta \geq 0$. A matrix $\mathbf{A} \in \mathbb{R}^{m \times n}$ is said to satisfy the $\mathop{S}$-REC$(\mathcal{S},\gamma,\delta)$ if $\forall \boldsymbol{x}_{1}, \boldsymbol{x}_{2} \in \mathcal{S},$
% \begin{equation}\label{eq11}
%     \Vert \mathbf{A}(\boldsymbol{x}_{1} - \boldsymbol{x}_{2}) \Vert_2 \geq \gamma \Vert \boldsymbol{x}_{1} - \boldsymbol{x}_{2} \Vert_2 - \delta.
% \end{equation}
% \end{definition}
% \noindent Classical CS algorithms impose REC~\citep{REC} on the measurement matrix $\mathbf{A}$ to recover an $s$-sparse signal. In GMCS recovery, $\mathop{S}$-REC generalizes REC to a set $\mathcal{S}$ of vectors that are in the range-space of the generator.\\
\indent Lemma~\ref{lemma_1} counts the number of $k$-dimensional faces or linear regions generated in $\mathbb{R}^{k}$ by simple arrangement of $h$ number of ($k-1$)-dimensional hyperplanes. The definition of {\it face} and {\it simple arrangement} is provided in the supplementary document.
\begin{lemma}\label{lemma_1}
~\citep{Bora} Let $h$ be the number of different ($k-1$)-dimensional hyperplanes. Each hyperplane partitions $\mathbb{R}^{k}$ such that all points inside a partition are on the same side. Then, the number of such partitions called as $k$-dimensional faces is $\mathcal{O}(h)^k$.

% activation $\mathcal{Z}\subseteq\mathbb{R}^{n}$ be a $k$-dimensional faces or linear regions and $f:\mathbb{R}\rightarrow \mathbb{R}$ be a piecewise-linear function with at most $t$ pieces. If $\mathcal{W} = \{\boldsymbol{y} \vert \boldsymbol{y} = f(\boldsymbol{x}), \forall\boldsymbol{x} \in \mathcal{Z}\}$, then $\mathcal{W}$ lies in the union of $\mathcal{O}((ht)^k)$ $k$-dimensional faces or linear regions.
\end{lemma}
\noindent For a given $d$-layer generator with $t$-piece activation function and at most $h$ nodes in each layer, from Lemma~\ref{lemma_1}, one can conclude that the number of $k$-dimensional faces or linear regions generated by a simple arrangement of $h$ hyperplanes in $\mathbb{R}^k$ equals $\mathcal{O}(ht^{kd})$~\citep{gajjar2020subspace}. The hyperplane is represented by the input point to the node for which the output is equal to zero.

\indent In the union-of-submanifolds signal model with sparsity $s$, the latent space $\mathcal{Z} \in \mathbb{R}^k$ is assumed to be composed of a union of $\binom{k}{s}$ number of $\mathbb{R}^s$ subspaces. We next provide a result (Lemma~\ref{lemma_2}), which gives the number of $s$-dimensional faces or linear regions generated by the simple arrangement of $h$ hyperplanes in $\mathbb{R}^{k}$. The proof of Lemma~\ref{lemma_2} is provided in the supplementary document.

\begin{lemma}\label{lemma_2}
Let $\mathcal{W}_i \subseteq \mathbb{R}^{k}$ be an $s$-dimensional subspace, where $i \in [1, 2, ..., \binom{k}{s}]$. Let $h$ be the number of hyperplanes in $\mathbb{R}^{s}$. Then the number of $s$-dimensional faces or linear regions generated by the simple arrangement of $h$ number of hyperplanes in $\mathbb{R}^{k}$ is $\mathcal{O}({(\frac{kh}{s})}^s)$.
\end{lemma}
We next consider a piecewise-linear activation function with $t$ pieces. 
If the transition points of the piecewise-linear function are fixed such that the input lies on either side of the transition, then for each set of inputs in a given partition, the output is a linear function of the input. From Lemma~\ref{lemma_2}, the number of $s$-dimensional faces or linear regions generated by the first layer of the generator $\mathbf{G}_{\boldsymbol{\theta}}$ is $\mathcal{O}({(\frac{kht}{s})}^s)$. The generator maps each $s$-dimensional face to a submanifold in $\mathbb{R}^n$.
Thus, the range-space of the generator $\mathbf{G}_{\boldsymbol{\theta}}(\mathbb{R}^k)$ with $d$ layers is the union of $\mathcal{O}\left({(\frac{kht}{s})}^{sd}\right)$ submanifolds in $\mathbb{R}^n$.

\begin{lemma}\label{lemma_3}
~\citep{Bora} Let $\mathcal{S} \subseteq \mathbb{R}^n$ be a submanifold, 
then $\mathbf{A} \in \mathbb{R}^{m\times n}$, with Gaussian i.i.d. entries $\mathbf{A}_{i,j}\sim \mathcal{N}(0,\frac{1}{m})$,
% with $\mathbf{A}_{i,j}\sim \mathcal{N}(0, \frac{1}{m})$
satisfies $\mathop{S}$-REC$(\mathcal{S},1 - \alpha,\delta)$ with probability $1-\exp{\left(-\Omega(\alpha^2m)\right)}$, if $m = \Omega(k/\alpha^2)$.
\end{lemma}
\noindent Lemma~\ref{lemma_3} states that, as long as there are enough linear measurements, the relative distance between the signals in the measurement domain is preserved up to an order $\delta$. Therefore, in SDLSS framework with $\mathcal{O}\left({(\frac{kht}{s})}^{sd}\right)$ submanifolds in $\mathbb{R}^n$ applying union bound over $\mathcal{O}({(\frac{kht}{s})}^{sd})$ submanifolds yields Lemma~\ref{lemma_4}.
The proof is provided in the supplementary document.
\begin{lemma}\label{lemma_4}
Let $\mathbf{G}_{\boldsymbol{\theta}}: \mathbb{R}^{k} \rightarrow \mathbb{R}^{n}$ be a generative model consisting of a $d$-layered neural network having piecewise-linear activation with $t$ pieces. If $h$ be the number of nodes per layer and $s$ be the sparsity assumed on the input latent space. Then, for a given value of $m$:
\begin{equation}\label{eq80}
    m = \Omega\left(\frac{1}{\alpha^2} sd\log \frac{kht}{s}\right),
\end{equation}
where $\alpha < 1$, the matrix $\mathbf{A}\in \mathbb{R}^{m \times n}$, with Gaussian i.i.d. entries $\mathbf{A}_{i,j}\sim \mathcal{N}(0,\frac{1}{m})$, satisfies $\mathop{S}$-REC$(\mathcal{S}_{s,\mathbf{G}_{\boldsymbol{\theta}}},(1-\alpha),0)$ with probability $1-\exp{\left(-\Omega(\alpha^2m)\right)}$.
\end{lemma}

\noindent The sample complexity bound for the SDLSS model containing a generator $\mathbf{G}_{\boldsymbol{\theta}}$ having a piecewise-linear activation function with $t$ pieces is given in Theorem~\ref{Thm}, which places a bound on the reconstruction error when the matrix satisfies $\mathop{S}$-REC on the union-of-submanifolds (represented by  $\mathcal{S}_{s,\mathbf{G}_{\boldsymbol{\theta}}}$). The proof is along the lines provided by~\citep{Bora}.

\begin{theorem}\label{Thm}
Let $\mathbf{G}_{\boldsymbol{\theta}}: \mathbb{R}^{k} \rightarrow \mathbb{R}^{n}$ be a generative model with $d$-layers having a piecewise-linear activation function with $t$ pieces and at most $h$ nodes in each layer. Let $\mathbf{A}\in \mathbb{R}^{m \times n}$ be a Gaussian matrix with i.i.d. entries $\mathbf{A}_{i,j}\sim \mathcal{N}(0,\frac{1}{m})$ that satisfies $\mathop{S}$-REC$(\mathcal{S}_{s,\mathbf{G}_{\boldsymbol{\theta}}},(1-\alpha),0)$. For any $\boldsymbol{x}\in \mathbb{R}^{n}$, if the number of measurements given by $\boldsymbol{y} = \mathbf{A}\boldsymbol{x} \in \mathbb{R}^m$ is $\mathcal{O}(sd\log \frac{kht}{s})$, where $s$ is the sparsity of the latent variable $\boldsymbol{z}$, then, with probability $1-\exp{\left(-\Omega(m)\right)}$, the following result holds:
\begin{equation}\label{eq12}
  \Vert \mathbf{G}_{\boldsymbol{\theta}}(\hat{\boldsymbol{z}}) - \boldsymbol{x} \Vert_2 \leq C \min_{\boldsymbol{z}\in \mathbb{R}^{k},\Vert \boldsymbol{z} \Vert_0 \leq s} \Vert \mathbf{G}_{\boldsymbol{\theta}}(\boldsymbol{z}) - \boldsymbol{x} \Vert_2 + 2 \epsilon,
\end{equation}
where $\mathbf{\hat{z}}$ minimizes $\Vert \boldsymbol{y} -\mathbf{A}\mathbf{G}_{\boldsymbol{\theta}}(\boldsymbol{z})\Vert_2$ to within additive $\epsilon$ of the optimum,  $C$ is a constant, $\Omega$ is the asymptotic lower bound, and $0 < \alpha < 1$.
\end{theorem}
\noindent The following result is a generalization of the sample complexity result by~\citep{Bora} for a generator model $\mathbf{G}_{\boldsymbol{\theta}}$ with ReLU activation (piecewise-linear with two pieces) and without the sparsity assumption on the latent space such that with $k = s$.
% gives the result by~\citep{Bora}. Therefore, a generative model considered in Theorem~\ref{Thm} with ReLU activation (piecewise-linear with two pieces), the sample complexity is $\mathcal{O}(sd\log \frac{kh}{s})$.

\begin{algorithm}[t]
   \caption{Sparsity-Driven Latent Space Sampling (SDLSS)}
   \label{algSDLSS}
\begin{algorithmic}
   \STATE {\bfseries Input:} Data batch = ${\{\boldsymbol{x}_i\}}_{i=1}^N$, sensing operator $\mathbf{A}\in \mathbb{R}^{m\times n}$ (linear) or $\mathbf{A}_{\boldsymbol{\phi}}$ (nonlinear), generator $\mathbf{G}_{\boldsymbol{\theta}}$, learning rate $\alpha$ and $\beta$, latent optimization steps $T$, batch size $N$, and  sparsity factor $s$.
   \REPEAT
   \STATE Initialize the generator and measurement network parameter $\boldsymbol{\theta}$ and $\boldsymbol{\phi}$,respectively.
   \FOR{$i=1$ {\bfseries to} $N$}
   \STATE Measure the signal $\boldsymbol{y}_i\leftarrow{} \mathbf{A}\boldsymbol{x}_i \text{ or }  \mathbf{A}_{\boldsymbol{\phi}}(\boldsymbol{x}_i$)
   \STATE Sample $\boldsymbol{z}\sim$ $\mathcal{N}(0,I)$
    \FOR{$t=1$ {\bfseries to} $T$}
    %\STATE Optimise  $\mathbf{\hat{z}}_i = \mathcal{P}_{\lambda \alpha}\left(\mathbf{\hat{z}}_i - \alpha  \nabla_{\hat{\boldsymbol{z}}} f(\boldsymbol{y}_i,\mathbf{\hat{z}}_i) \right)$
    \STATE $\hat{\boldsymbol{z}}_i = \mathcal{P}_{s}\left(\mathbf{{z}}_i - \beta  \nabla_{{\boldsymbol{z}}} f(\boldsymbol{y}_i,\mathbf{{z}}_i) \right)$
    \ENDFOR
   \ENDFOR

  \STATE Update 1:$\boldsymbol{\theta} \xleftarrow{} \boldsymbol{\theta} - \alpha \frac{\partial{\mathcal{L}}}{\partial \boldsymbol{\theta}}$ \text{ } \text{(linear)}
  \STATE \text{ or } 
  \STATE Update 2:$\boldsymbol{\theta} \xleftarrow{} \boldsymbol{\theta} - \alpha \frac{\partial{\mathcal{L}}}{\partial \boldsymbol{\theta}}, \text{ }\boldsymbol{\phi} \xleftarrow{} \boldsymbol{\phi} - \alpha \frac{\partial{\mathcal{L}}}{\partial \boldsymbol{\phi}}$\text{ } \text{(nonlinear)}

   \UNTIL{$\Vert \boldsymbol{y} - \mathbf{A} \mathbf{G}_{\boldsymbol{\theta}}(\boldsymbol{z}) \Vert_2 \leq \epsilon$ \text{ or }  $\Vert \boldsymbol{y} - \mathbf{A}_{\boldsymbol{\phi}} \mathbf{G}_{\boldsymbol{\theta}}(\boldsymbol{z}) \Vert_2 \leq \epsilon$}
\end{algorithmic}
\label{SDGPAlgo}
\end{algorithm}

\section{Proximal Meta-learning}
Meta-learning or {\it learning-to-learn} proposed by~\citep{finn} is a novel training strategy that has been shown to improve the performance of models involving multiple tasks. Meta-learning can solve new tasks efficiently using only a small number of training samples. Deep compressed sensing (DCS)~\citep{Yan} employed meta-learning to jointly optimize the latent variable $\boldsymbol{z}$ and the parameters $\boldsymbol{\theta}$ and $\boldsymbol{\phi}$ of the generative and measurement models, respectively, in the nonlinear settings and demonstrated that the meta-learning results in faster reconstruction from fewer measurements.
We propose a proximal meta-learning (PML) algorithm that fuses the idea of proximal gradient-descent together with meta-learning to promote sparsity in the latent space while optimizing the cost in Eq.~\eqref{EQ:LA}.

\indent In PML algorithm, the sparse coding of each data sample in a batch is considered as a task $\mathcal{T}_i$ and the corresponding latent variable $\boldsymbol{z}_i$ (such that $\boldsymbol{x}_i = \mathbf{G}_{\boldsymbol{\theta}}(\boldsymbol{z}_i)$ and $\boldsymbol{y}_i = \mathbf{A}\boldsymbol{x}_i)$) as the parameter associated with task $\mathcal{T}_i$. The generator loss $\mathcal{L}_\mathbf{G}$ associated with all the tasks is given as:
\begin{equation}\label{19}
    \mathcal{L}_\mathbf{G} = \frac{1}{N} \sum_{i=1}^N  \left\{\Vert \boldsymbol{y}_i - \mathbf{A}\mathbf{G}_{\boldsymbol{\theta}}(\boldsymbol{z}_i) \Vert_2 + \Vert\boldsymbol{z}_i\Vert_0 \right\},
\end{equation}
where $N$ is the batch-size and $\Vert \cdot\Vert_0$ denotes the $\ell_0$-pseudo-norm.
The latent space is optimized to adapt to each task $\mathcal{T}_i$ with a few (up to five) proximal gradient-descent updates on the latent variable $\boldsymbol{z}_i$:
\begin{align*}\label{eq21}
\boldsymbol{\hat{z}}_i & = \argminA_{\boldsymbol{z}_i} \left\{f(\boldsymbol{y}_i,\boldsymbol{z}_i) = \Vert \boldsymbol{y}_i - \mathbf{A}\mathbf{G}_{\boldsymbol{\theta}}(\boldsymbol{z}_i) \Vert_2\right\},
\\ & = \mathcal{P}_{s}\left(\boldsymbol{z}_i - \beta  \nabla_{{\boldsymbol{z}}} f(\boldsymbol{y}_i,\boldsymbol{z}_i) \right),
% 	\mathbf{\hat{z}}_i \xleftarrow{} \boldsymbol{z} - \alpha \Delta_{\boldsymbol{\theta}} \mathcal{L_T}_i(f_{\boldsymbol{\theta}}),
\end{align*}
where $\beta$ is the learning rate, and $ \mathcal{P}_{s}$ is the hard-thesholding operator that sets all but the largest (in magnitude) $s$ elements of its argument to 0.

\indent The model parameter $\boldsymbol{\theta}$ is trained by optimizing the measurement loss $\mathcal{L}_\mathbf{G}$ and $\mathop{S}$-REC loss $\mathcal{L}_\mathbf{A}$ across all the tasks $\mathcal{T}_i$. The parameter updates are performed via stochastic gradient-descent (SGD)~\citep{robbins1951stochastic}:
\begin{equation}
\boldsymbol{\theta} \xleftarrow{} \boldsymbol{\theta} -  \alpha\frac{\partial{(\mathcal{L}_\mathbf{G}} + \mathcal{L}_\mathbf{A})}{\partial \boldsymbol{\theta}},
% \quad 
% \boldsymbol{\phi} \xleftarrow{} \boldsymbol{\phi} -  \alpha\frac{\partial{(\mathcal{L}_\mathbf{G}} + \mathcal{L}_\mathbf{A})}{\partial \boldsymbol{\phi}},
\label{metastep}
% 	{\boldsymbol{\theta}} \xleftarrow{} {\boldsymbol{\theta}} - \beta\Delta_\theta {\sum_{\mathcal{T}_i}{\mathcal{L_T}_i(f_{\boldsymbol{\theta} - \alpha {\Delta_{\boldsymbol{\theta}} \mathcal{L_T}_i(f_{\hat{\boldsymbol{\theta}}})}}})},
\end{equation}
where $\alpha$ is the corresponding meta step-size parameter. The SDLSS approach is summarized in Algorithm~\ref{algSDLSS}.

\section{Experimental Results}
We perform experimental validation on standard datasets such as Fashion-MNIST~\citep{fashionmnist}, CIFAR-$10$~\citep{CIFAR} and CelebA~\citep{celeba}, with the dimensionality of the images in the datasets being $28\times28$, $32\times32$ and $64\times64$, respectively. The generator is a two-layered feed forward neural network for Fashion-MNIST dataset with $500$ neurons in each hidden layer and leaky ReLU as the activation function. In contrast, a standard DCGAN generator~\citep{DCGAN} is used for CIFAR-$10$ and CelebA datasets. These settings are the same as those reported by~\citep{Yan} and have been chosen to enable a fair comparison. The efficacy of the proposed SDLSS model is demonstrated in comparison with DCS~\citep{Yan} in terms of three objective metrics: peak signal-to-noise ratio (PSNR = $-10 \log (\Vert \boldsymbol{x} - \mathbf{G}_{\boldsymbol{\theta}}(\boldsymbol{z})\Vert_2^2)$ dB), with peak value considered to be unity; per-pixel reconstruction error (RE $= 10 \log(\Vert \boldsymbol{x} - \mathbf{G}_{\boldsymbol{\theta}}(\boldsymbol{z})\Vert_2^2/n)$ dB), where $n$ is the total number of pixels; and the structural similarity index metric (SSIM)~\citep{ssim}. The metrics are averaged over all the test images of a batch. A batch size of $64$ images is considered for all the datasets.

\indent We would first like to highlight the superiority of a learned nonlinear sensing operator in comparison to the linear one. We generate two sets of measurements $\boldsymbol{y}$ from the given training samples $\boldsymbol{x}$, one corresponding to a linear operator and the other corresponding to a nonlinear operator. The linear operator $\mathbf{A}\in \mathbb{R}^{m \times n}$ is a random Gaussian matrix with i.i.d. entries such that $\mathbf{A}_{i,j}\sim \mathcal{N}(0.\frac{1}{m})$. The nonlinear sensing operator is either a feedforward neural network (for Fashion-MNIST) or a convolutional neural network (for CelebA and CIFAR-10).
% The parameters of the generator $\mathbf{G}_{\boldsymbol{\theta}}$ and the sensing operator $\mathbf{A}_{\boldsymbol{\phi}}$ are updated as per Eq.~\eqref{metastep} at the end of training over a batch, while the latent space is updated in three steps of gradient-descent for each of the samples in the training batch.
\begin{figure}[t]
  \centering
  \begin{subfigure}{.4\textwidth}
    \centering\includegraphics[width=\textwidth]{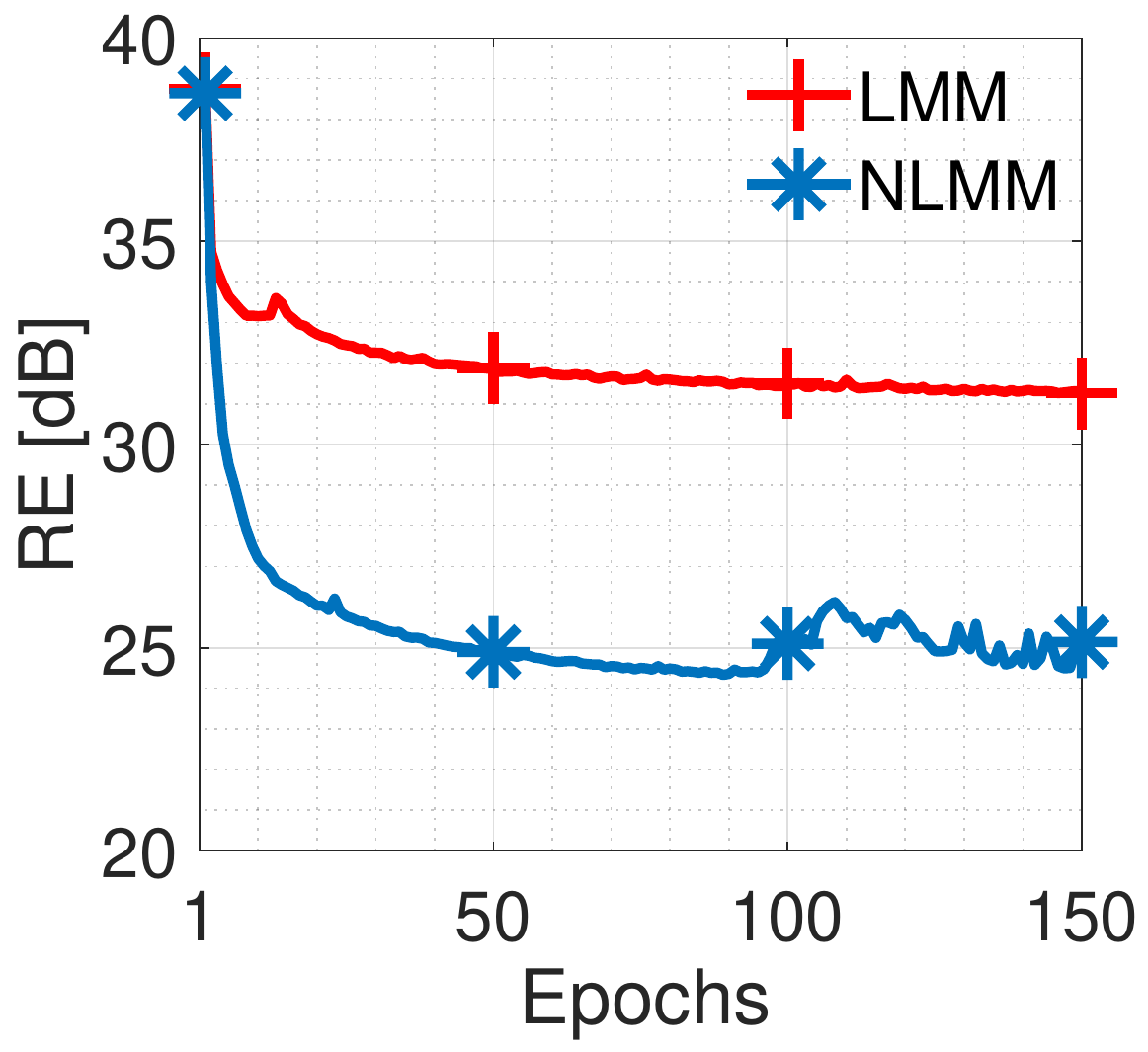}
    \caption{}
    \label{RE_CIFAR_Tr}
  \end{subfigure}
  \begin{subfigure}{.4\textwidth}
    \centering\includegraphics[width=\textwidth]{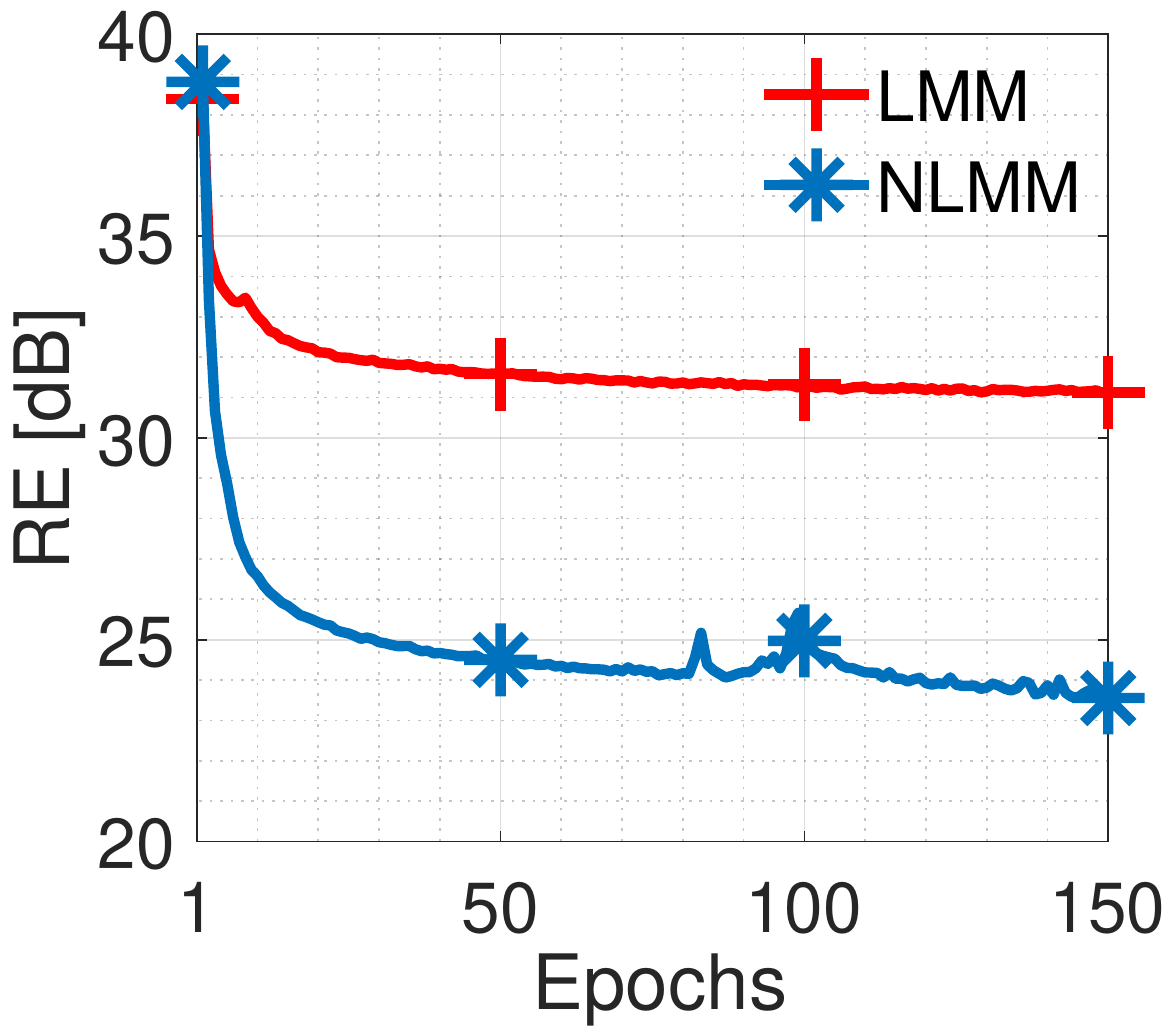}
    \caption{}
    \label{RE_MNIST_Tr}
  \end{subfigure}
  \caption{Validation error (Fashion-MNIST) as a function of the number of epochs for a given number of measurements $m$ = $5$: (a) $k$ = $100$, (b) $k$ = $784$. LMM stands for the linear measurement model and NLMM corresponds to the nonlinear one.}
  \label{NonlinLin}
\end{figure}

\begin{figure}[t]
  \centering
  \begin{subfigure}{.8\textwidth}
    \centering\includegraphics[width=\textwidth]{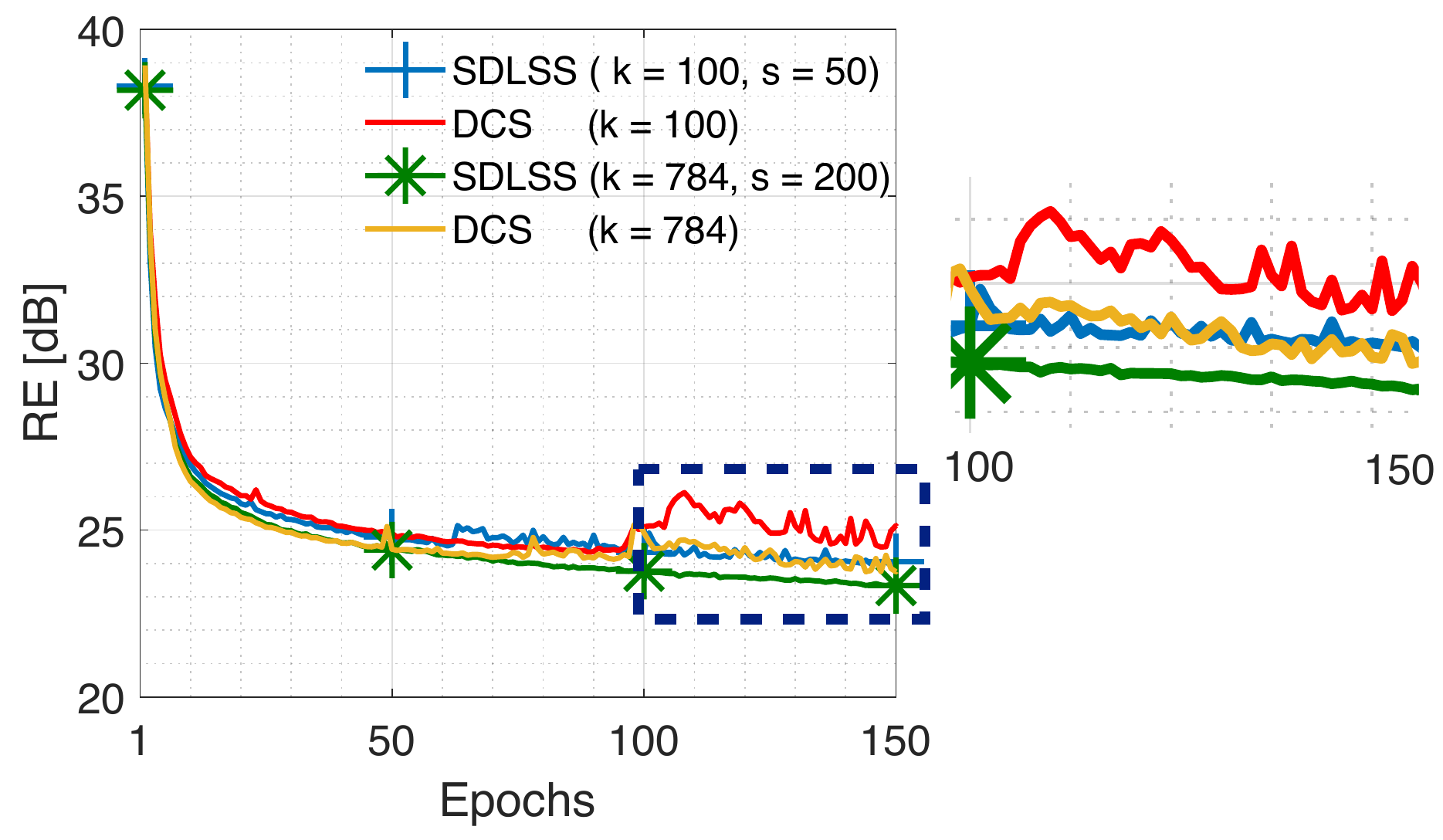}
    % \caption{PSNR [dB] for $k = 100$ and $s = 80$.}
    \label{RE_CIFAR_k100}
  \end{subfigure}
  \caption{Fashion-MNIST: Validation error as function of epochs for a given measurement $m$ = $5$.}
  \label{Nonlin784100}
\end{figure}
\begin{figure}[t]
  \centering
  \begin{subfigure}{.6\textwidth}
    \centering\includegraphics[width=\textwidth]{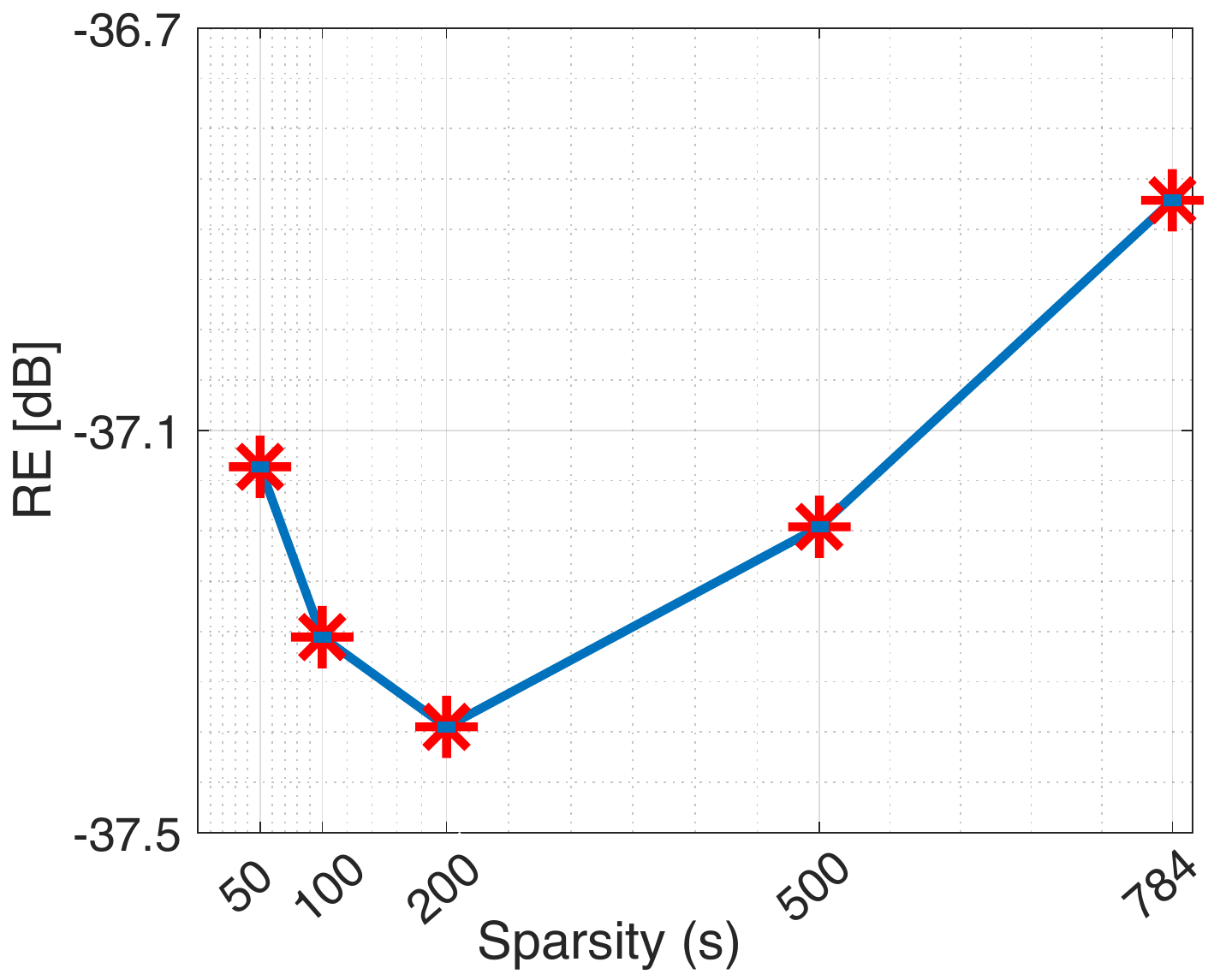}
    % \caption{PSNR [dB] for $k = 100$ and $s = 80$.}
    \label{RE_CIFAR_k784}
  \end{subfigure}
  \caption{Fashion-MNIST: Reconstruction error on test data as a function of sparsity for a given measurements $m = 10$ and latent dimension $k = 784$.}
  \label{RE_sparse}
\end{figure}
\begin{figure*}[t]
\centering
\begin{subfigure}{.32\textwidth}
    \centering\includegraphics[width=\textwidth]{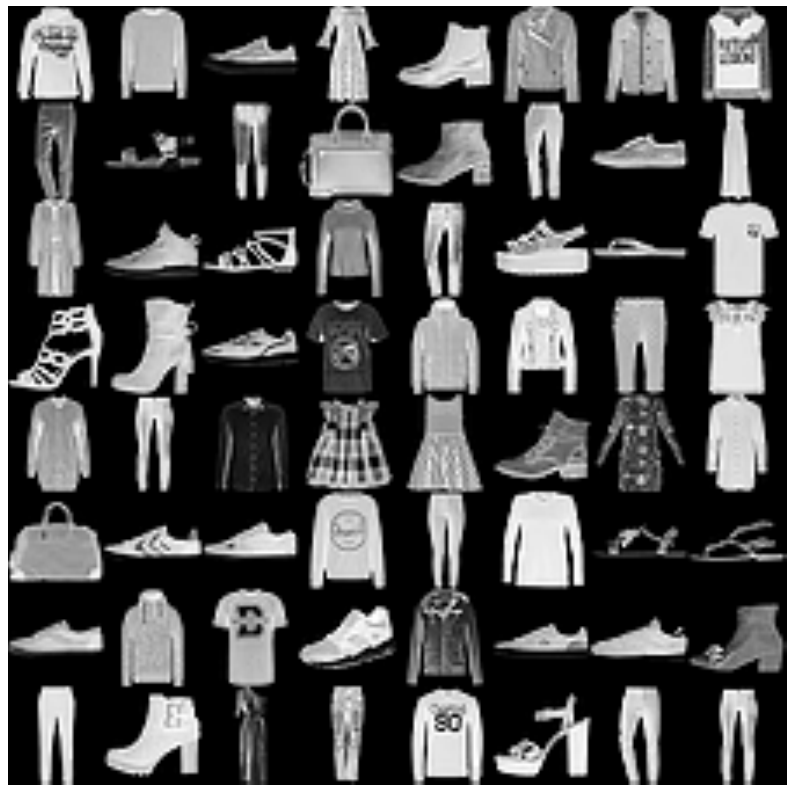}
    \caption*{Ground truth}
  \end{subfigure}
  ~
\begin{subfigure}{.32\textwidth}
    \centering\includegraphics[width=\textwidth]{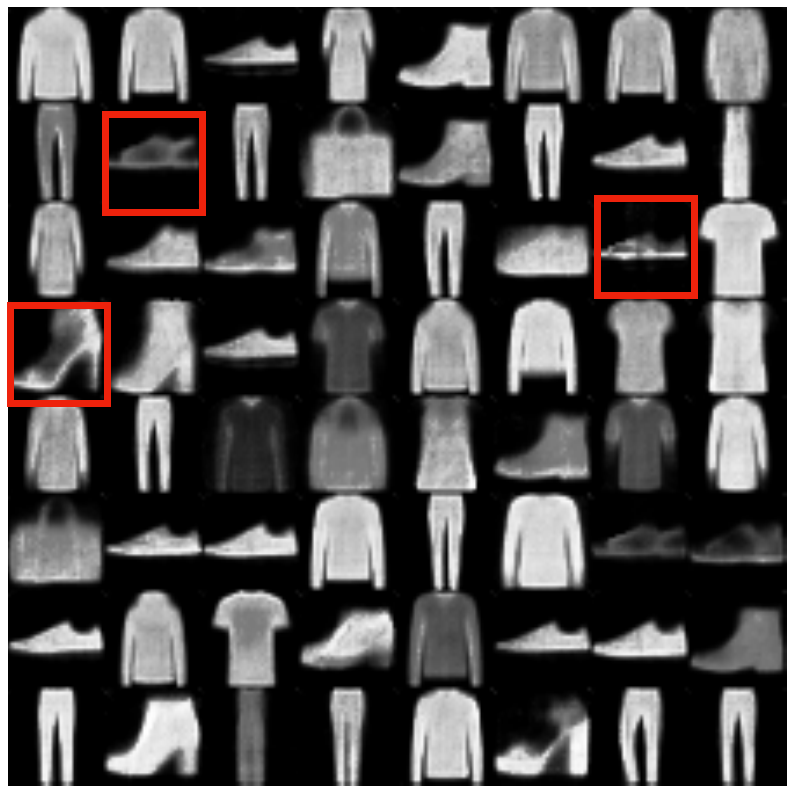}
    \caption*{DCS, PSNR $=17.58$ dB,  SSIM $=0.7147$}
  \end{subfigure}
  ~
\begin{subfigure}{.32\textwidth}
    \centering\includegraphics[width=\textwidth]{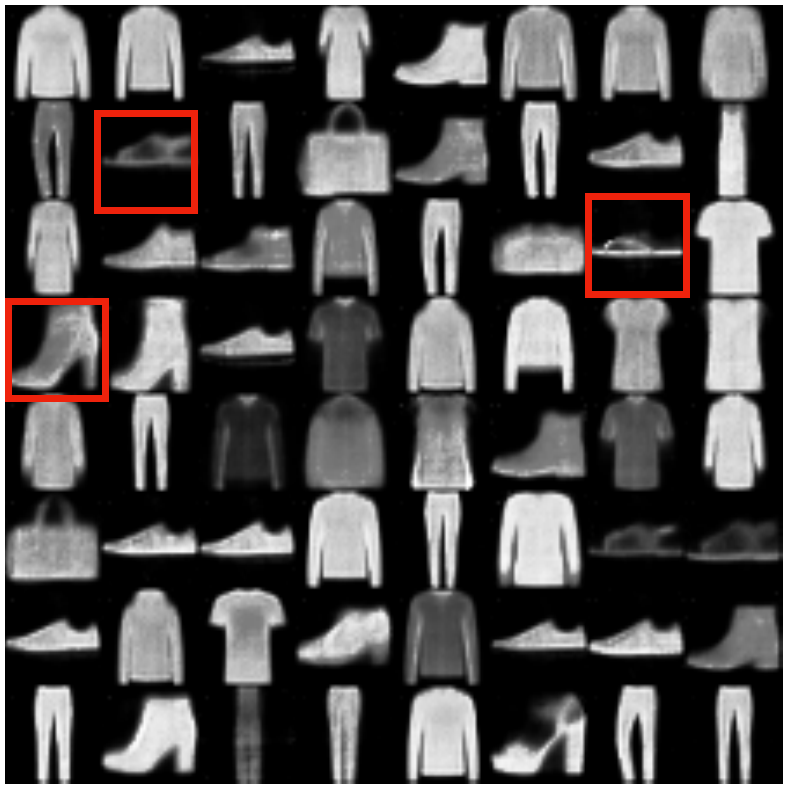}
    \caption*{SDLSS, PSNR $=\mathbf{17.78}$ dB,  SSIM $=\mathbf{0.7312}$}
  \end{subfigure}
 
\caption{Reconstructed images (Fashion-MNIST dataset) with corresponding PSNR/SSIM values for $m = 5$ and $k = 784$. The sparsity factor $s = 200$ in SDLSS. The SDLSS framework gives images with sharper features (illustrated in red boxes).}
\label{RE_FMNIST}
\end{figure*}
\begin{figure*}[t]
\centering
    \includegraphics[width=\textwidth]{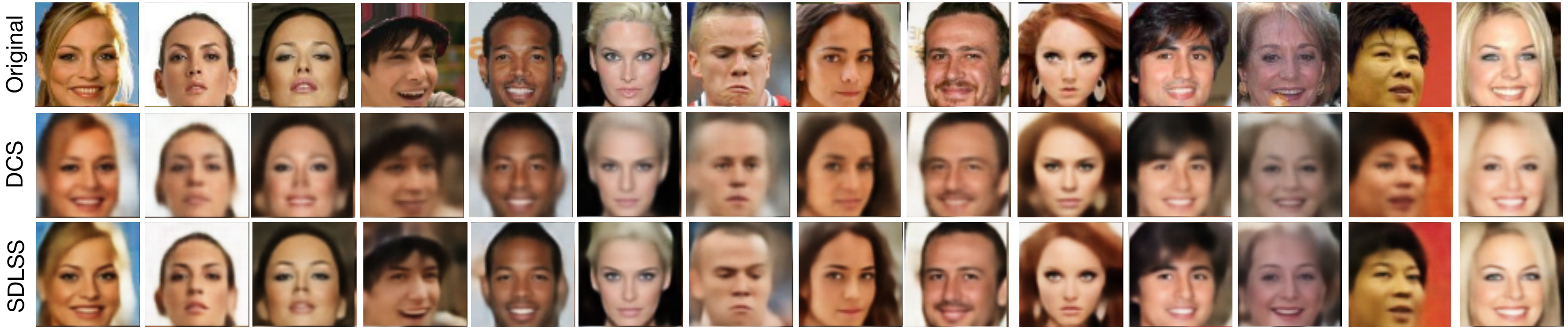}
\caption{Performance illustration on CelebA dataset. Row one: ground-truth image; Rows two and three: images reconstructed by DCS and SDLSS, respectively. $m = 200$ and $k = 4096$. The sparsity factor $s$ in SDLSS is set to $2000$.}
\label{RE_CelebA}
\end{figure*}

In order to assess and compare the performance of the SDLSS framework with a linear or a learned nonlinear measurement model, we conduct experiments on Fashion-MNIST dataset with varying values of the dimension of the latent space. 
The maximum value of the dimension of the latent space is set exactly equal to the dimension of the generator output for the FashionMINIST dataset, i.e., $28 \times 28 = 784$). On the other hand, for CIFAR-$10$, the maximum value of $k$ is set to $32 \times 32 = 1024$, and for CelebA, it is set to $64 \times 64 = 4096$, i.e., the dimension of a single channel in the image. The plot of the reconstruction error for two choices of the latent space dimension is shown in Fig.~\ref{NonlinLin}. The figure shows that the performance of the learned nonlinear model is superior to the linear one. Henceforth, we consider the learned nonlinear measurement model for comparison with DCS~\citep{Yan}.\\
\indent The next experiment demonstrates the impact of sparsity on the reconstruction error. We consider two cases with and without sparsity (i.e., DCS) considering two choices of the latent space dimension: $k$ = $100$ (latent space dimension less than that of the input) and $k$ = $784$ (latent space dimension equal to that of the input). The reconstruction error is shown in Fig.~\ref{Nonlin784100}. From the figure, we observe that the reconstruction error in case of DCS decreases with increase in the dimensionality of the latent space, which is consistent with the observation reported by \citep{invetgan}. The reconstruction error behavior in case of SDLSS is also similar. However, the decrease in reconstruction error of DCS is minimal compared to that of SDLSS. Further, the performance of SDLSS framework with sparsity $s = 50$ and latent dimension $k = 100$ is on par with the DCS framework with latent dimension $k = 784$. 
Fig.~\ref{RE_sparse} shows the reconstruction error of SDLSS as a function of the sparsity level $s$ for $k=784$. We observe that there is an optimal sparsity level for which the reconstruction error is the least. Therefore, the effective latent space dimension is much smaller than $784$. This experiment underscores the importance of introducing sparsity into the generative model based CS framework.
% \indent We next demonstrate that sparsity enforced in the latent space {\it disentangles} various features of an image. 
Fig.~\ref{RE_FMNIST} shows the images reconstructed using both DCS and SDLSS frameworks in comparison with the ground truth. 
The PSNR and SSIM values indicated in Fig.~\ref{RE_FMNIST} are calculated for a given batch and these metrics are higher for the SDLSS technique than DCS. 
Visual inspection shows that the images generated by SDLSS have sharper features (highlighted using red boxes) compared with those generated by DCS. This experiment shows that introducing sparsity has the effect of minimizing overlap between features by suppressing redundancy of the features. In order to further confirm the finding, we carry out the same experiment on the CelebA dataset, which comprises color images. The colour is a new feature here and we observe from Fig.~\ref{RE_CelebA} that SDLSS preserves the color information more accurately than DCS. The SDLSS images are also sharper than DCS images.
The PSNR and SSIM values averaged over all test images in a batch with different values of $m$ for Fashion-MNIST, CIFAR-$10$, and CelebA datasets are computed and summarized in Tables~\ref{table_FMNIST}, ~\ref{table_CIFAR}, and ~\ref{table_CelebA}, respectively. The SDLSS outperforms DCS at lower compression rates over all the datasets under consideration.

\begin{table}[tbt]
\caption{Performance comparison between DCS and the SDLSS ({\bf ours}) approaches considering Fashion-MNIST dataset. The latent space dimension $k = 784 $.}
\label{table_FMNIST}
\vskip 0.15in
\begin{center}
\begin{small}
\begin{sc}
\resizebox{12cm}{!}
{\begin{tabular}{l|c|c|c}
\toprule
Method & Metric & $m = 5$ & $m = 10$ \\
\midrule
\midrule
DCS    & PSNR [dB] &$17.26$ $\pm$ $0.35$& $18.60$ $\pm$ $0.35$  \\
    & SSIM & $0.674$ $\pm$ $0.020$& $0.741$ $\pm$ $0.017$  \\
    \hline
SDLSS & PSNR [dB]& $\mathbf{17.31}$ $\pm$ $\mathbf{0.37}$& $\mathbf{18.71}$ $\pm$ $\mathbf{0.35}$ \\
 ($s$=200)& SSIM & $\mathbf{0.691}$ $\pm$ $\mathbf{0.019}$& $\mathbf{0.755}$ $\pm$ $\mathbf{0.017}$ \\
% DCS(R)    & 0.0126 & 0.0114&  \\

\bottomrule
\end{tabular}
}
\end{sc}
\end{small}
\end{center}
\vskip -0.1in
% \label{CT}
\end{table}

\begin{table}[t]
\caption{Performance comparison between DCS and SDLSS approaches on CIFAR-$10$ dataset ($k = 1024$).}
\label{table_CIFAR}
\vskip 0.15in
\begin{center}
\begin{small}
\begin{sc}
\resizebox{12cm}{!}{\begin{tabular}{l|c|c|c}
\toprule
Method & Metric & $m = 25$ & $m = 50$ \\
\midrule
\midrule
DCS    & PSNR [dB]&$19.97$ $\pm$ $0.25$& $21.90$ $\pm$ $0.25$  \\
    & SSIM & $0.611$ $\pm$ $0.017$& $\mathbf{0.721}$ $\pm$ $\mathbf{0.013}$  \\
    \hline
SDLSS & PSNR [dB]& $\mathbf{20.36}$ $\pm$ $\mathbf{0.23}$& $\mathbf{22.22}$ $\pm$ $\mathbf{0.25}$ \\
 ($s$=500)& SSIM & $\mathbf{0.611}$ $\pm$ $\mathbf{0.017}$& $\mathbf{0.718}$ $\pm$ $\mathbf{0.014}$ \\
% DCS(R)    & 0.0126 & 0.0114&  \\

\bottomrule
\end{tabular}
}
\end{sc}
\end{small}
\end{center}
\vskip -0.1in
% \label{CT}
\end{table}

% \begin{table}[!ht]
% \caption{Performance comparison between DCS and SDLSS ({\bf ours}) on CelebA dataset ($k = 4096$).}
% \label{table_CelebA}
% \vskip 0.15in
% \begin{center}
% \begin{small}
% \begin{sc}
% \begin{tabular}{l|c|c|c}
% \toprule
% Method & Metric & $m = 50$ & $m = 100$ \\
% \midrule
% \midrule
% DCS & PSNR [dB] &$21.87$ $\pm$ $0.23$& $23.40$ $\pm$ $0.27$  \\
%     & SSIM & $0.811$ $\pm$ $0.011$& $0.852$ $\pm$ $0.017$  \\
%     \hline
% SDLSS & PSNR [dB] & $\mathbf{22.02}$ $\pm$ $\mathbf{0.24}$& $\mathbf{23.58}$ $\pm$ $\mathbf{0.26}$ \\
%  ($s$=500)& SSIM & $\mathbf{0.818}$ $\pm$ $\mathbf{0.011}$& $\mathbf{0.859}$ $\pm$ $\mathbf{0.008}$ \\
% % DCS(R)    & 0.0126 & 0.0114&  \\

% \bottomrule
% \end{tabular}
% \end{sc}
% \end{small}
% \end{center}
% \vskip -0.1in
% % \label{CT}
% \end{table}

\begin{table}[!ht]
\caption{Performance comparison between DCS and SDLSS ({\bf ours}) on CelebA dataset ($k = 4096$).}
\label{table_CelebA}
\vskip 0.15in
\begin{center}
\begin{small}
\begin{sc}
\resizebox{12cm}{!}{\begin{tabular}{l|c|c|c}
\toprule
Method & Metric & $m = 100$ & $m = 200$ \\
\midrule
\midrule
DCS & PSNR [dB] &$21.87$ $\pm$ $0.23$& $22.75$ $\pm$ $0.27$  \\
    & SSIM & $0.811$ $\pm$ $0.011$& $0.844$ $\pm$ $0.017$  \\
    \hline
SDLSS & PSNR [dB] & $\mathbf{23.90}$ $\pm$ $\mathbf{0.24}$& $\mathbf{25.76}$ $\pm$ $\mathbf{0.26}$ \\
 ($s$=2000)& SSIM & $\mathbf{0.865}$ $\pm$ $\mathbf{0.011}$& $\mathbf{0.901}$ $\pm$ $\mathbf{0.008}$ \\
% DCS(R)    & 0.0126 & 0.0114&  \\

\bottomrule
\end{tabular}}
\end{sc}
\end{small}
\end{center}
\vskip -0.1in
% \label{CT}
\end{table}

\section{Conclusions}
We addressed the compressed sensing problem using a deep generative prior and proposed a sparsity driven latent space sampling (SDLSS) framework and a union-of-submanifolds signal model to capture the data distribution.
We developed the proximal meta-learning (PML) algorithm by combining proximal gradient-descent and meta-learning for promoting sparsity while optimizing the latent space along with training of the generator model in the SDLSS framework.
We considered both linear and learned nonlinear sensing mechanisms and showed that the latter operating within the SDLSS framework is superior compared to DCS provided that the level of sparsity in the latent space is chosen appropriately.
In addition, we derived the sample complexity results for a linear measurements with the generator prior containing a neural network with piecewise linear activation function with atmost $t$-pieces and showed that the results derived by~\citep{Bora} are a special case.
The efficacy of the SDLSS over DCS was demonstrated with application to compressed sensing on standard datasets such as Fashion-MNIST, CIFAR-$10$, and CelebA.
The question of determining the optimal level of sparsity in the latent space remains an open one.

% \begin{contributions} % will be removed in pdf for initial submission,
%                       % so you can already fill it to test with the
%                       % ‘accepted’ class option
%     Briefly list author contributions.
%     This is a nice way of making clear who did what and to give proper credit.

%     H.~Q.~Bovik conceived the idea and wrote the paper.
%     Coauthor One created the code.
%     Coauthor Two created the figures.
% \end{contributions}

% \begin{acknowledgements} % will be removed in pdf for initial submission,
%                          % so you can already fill it to test with the
%                          % ‘accepted’ class option
%     Briefly acknowledge people and organizations here.

%     \emph{All} acknowledgements go in this section.
% \end{acknowledgements}

% \bibliography{uai2021-template}
\newpage
\appendix
% NOTE: necessary when ptmx or no mathfont class option is given
\providecommand{\upGamma}{\Gamma}
\providecommand{\uppi}{\pi}
% \section*{Supplementary Document}
\title{Supplementary Document}
\maketitle
\noindent We review the definitions of {\it simple arrange} and {\it face} provided in~\citep{LecGeom} and use them to prove Lemma~\ref{lemma_2} and Lemma~\ref{lemma_4} presented in the main paper.

% \begin{definition}\label{SREC}
% (Bora \it {et al.}~\citep{Bora}) Let $\mathcal{S} \subseteq \mathbb{R}^{n}$, $\gamma > 0$, and $\delta \geq 0$. A random matrix $\mathbf{A} \in \mathbb{R}^{m \times n}$ is said to satisfy the $\mathop{S}$-REC$(\mathcal{S},\gamma,\delta)$ if $\forall \boldsymbol{x}_{1}, \boldsymbol{x}_{2} \in \mathcal{S},$
% \begin{equation}\label{eq11}
%     \Vert \mathbf{A}(\boldsymbol{x}_{1} - \boldsymbol{x}_{2}) \Vert_2 \geq \gamma \Vert \boldsymbol{x}_{1} - \boldsymbol{x}_{2} \Vert_2 - \delta.
% \end{equation}
% \end{definition}

\begin{definition}\label{simple}
(~\citep{LecGeom})Let $\mathcal{H}$ be a set of hyperplanes in $\mathbb{R}^d$. If the intersection of every $k$ hyperplane is $(d-k$)-dimensional, then the arrangement of $\mathcal{H}$ is called simple.
\end{definition}

\begin{definition}\label{linear_region}
(~\citep{LecGeom})Let $\mathcal{H}$ be a set of hyperplanes in $\mathbb{R}^d$. A simple arrangement of a finite set $\mathcal{H}$ of hyperplanes partitions $\mathbb{R}^d$ into open convex sets with dimensions ranging from $0$ to $d$. A $d$-dimensional open convex set is called as $d$-dimensional face or $d$-dimensional linear region.
\end{definition}

\section{Proof of Lemma 2}
\begin{proof}
The proof is based on induction and is provided for the case of ReLU activation (piecewise linear with two piece) and $k=s$ in Lemma \ref{lemma_1}~\citep{Bora}. For $k=s$ the number of $k$-dimensional faces or linear regions generated by $h$ number of hyperplanes in $\mathbb{R}^{k}$ is $\mathcal{O}(h^k)$.

For $k \neq s$ and $k \gg s$, where $s$ is the sparsity of the latent dimension, the number of $s$-dimensional subspaces $\mathcal{W}_i \subseteq \mathbb{R}^{k}$ is $\mathcal{O}({(\frac{k}{s})}^s)$ (upper bound on $\binom{k}{s}$). According to Lemma \ref{lemma_1}, $h$ hyperplanes divide $\mathcal{W}_i \in \mathbb{R}^s$ into $\mathcal{O}(h^s)$ number of $s$-dimensional faces or linear regions. Each $\mathcal{W}_i \in \mathbb{R}^s$ is partitioned independently by the hyperplanes. Therefore, the total the number of $s$-dimensional faces or  linear regions in $\mathbb{R}^k$ is $\mathcal{O}({(\frac{kh}{s})}^s)$.
\end{proof}
\section{Proof of Lemma 4}
\begin{proof}
Let $s$ be the support of the input latent space and $h$ be the number of nodes in the first layer of the generator network. Each node can be considered as a hyperplane in $\mathbb{R}^{k}$. Therefore, from Lemma~\ref{lemma_1}, Lemma~\ref{lemma_2} and the result for piecewise linear activation function with atmost $t$ piece presented by~\citep{gajjar2020subspace}, the number of $s$-dimensional faces or linear regions in $\mathbb{R}^k$ generated by the first layer of the generator $\mathbf{G}_{\boldsymbol{\theta}}$ with $t$-piece activation is $\mathcal{O}({(\frac{kht}{s})}^s)$.
 Since there are $d$ layers in the generator network, the input space $\mathbb{R}^k$ is partitioned into at most $\mathcal{O}({(\frac{kht}{s})}^{sd})$ number of $s$-dimensional faces or linear regions. 
The generator $\mathbf{G}_{\boldsymbol{\theta}}$ maps each $s$-dimensional faces or linear region to a corresponding submanifold.
Thus, the range-space of the generator $\mathbf{G}_{\boldsymbol{\theta}}(\mathbb{R}^k)$ with $d$ layers is the union of $\mathcal{O}({(\frac{kht}{s})}^{sd})$ number of submanifolds.

% $s$-dimensional faces or linear regions in $\mathbb{R}^n$, which implies union of $\binom{k}{s}$ number of $s$-dimensional submanifolds given as $\mathbf{G}_{\boldsymbol{\theta}}(\mathbb{R}^{k}) =  \mathcal{S}_{s,\mathbf{G}_{\boldsymbol{\theta}}}$.
% The union of  $\mathcal{O}({(\frac{kc}{s})}^{sd})$ $s$-dimensional subspaces forms the range-space on the generator network $\mathbf{G}_{\boldsymbol{\theta}}$ given as $\mathbf{G}_{\boldsymbol{\theta}}(\mathbb{R}^{k}) =  \mathcal{S}_{s,\mathbf{G}_{\boldsymbol{\theta}}}$.
%  The generator maps each $s$-dimensional subspace to an $s$-dimensional submanifold in $\mathbb{R}^n$.

From Lemma \ref{lemma_3}, applying union bound over ${(\frac{kht}{s})}^{sd}$ number of submanifolds implies that  $\mathbf{A}$ satisfies the $\mathop{S}$-REC$(\mathcal{S}_{s,\mathbf{G}_{\boldsymbol{\theta}}},(1-\alpha),0)$ with probability $\displaystyle 1-{\left(\frac{kht}{s}\right)}^{sd} \exp^{-\Omega(\alpha^2m)}$.
Thus, $\mathbf{A}$ satisfies the $\mathop{S}$-REC$(\mathcal{S}_{s,\mathbf{G}_{\boldsymbol{\theta}}},(1-\alpha),0)$ with probability $1-\exp^{-\Omega(\alpha^2m)}$, if $m$ is bounded as given below:
\begin{equation}\label{Aeq9}
    m = \Omega\Big(\frac{1}{\alpha^2} sd\log \frac{kht}{s}\Big).
\end{equation}

\end{proof}

\bibliographystyle{unsrtnat}
\bibliography{references}  %%% Uncomment this line and comment out the ``thebibliography'' section below to use the external .bib file (using bibtex) .





% \end{thebibliography}

\end{document}